\pdfoutput=1
\documentclass[11pt]{article}

\usepackage{acl}

\usepackage{times}
\usepackage{latexsym}

\usepackage[T1]{fontenc}
\usepackage[utf8]{inputenc}

\usepackage{microtype}

\usepackage{inconsolata}

\usepackage{graphicx}
\usepackage{color}
\usepackage{subfig}
\usepackage{multirow}
\usepackage{booktabs}
\usepackage{amsmath}

\title{FOCUS: Forging Originality through Contrastive Use in Self-Plagiarism for Language Models}

\author{
        Kaixin Lan$^1$\thanks{~~Equal Contribution}~~~~
        Tao Fang$^{1*}$~~~~
        Derek F. Wong$^1\thanks{~~Corresponding Author}$~~~~
        Yabo Xu$^2$~~~~ \\
        \textbf{Lidia S. Chao$^1$}~~~~
        \textbf{Cecilia G. Zhao$^3$}~~~~
       \\
    $^1$NLP$^2$CT Lab, Department of Computer and Information Science, University of Macau \\
    \texttt{nlp2ct.\{Kaixin,taofang\}@gmail.com, \{derekfw,lidiasc\}@um.edu.mo} \\
     $^2$Guangdong Hengqin DataStory Information Technology Ltd.\\
       \texttt{arber@datastory.com.cn} \\
    $^3$Department of English, Faculty of Arts and Humanities, University of Macau\\
      \texttt{czhao@um.edu.mo}  
    }

\begin{document}
\maketitle

\begin{abstract}
Pre-trained Language Models (PLMs) have shown impressive results in various Natural Language Generation (NLG) tasks, such as powering chatbots and generating stories. However, an ethical concern arises due to their potential to produce verbatim copies of paragraphs from their training data. This is problematic as PLMs are trained on corpora constructed by human authors. As such, there is a pressing need for research to promote the generation of original content by these models. In this study, we introduce a unique ``self-plagiarism'' contrastive decoding strategy, aimed at boosting the originality of text produced by PLMs. Our method entails modifying prompts in LLMs to develop an amateur model and a professional model. Specifically, the amateur model is urged to plagiarize using three plagiarism templates we have designed, while the professional model maintains its standard language model status. This strategy employs prompts to stimulate the model's capacity to identify non-original candidate token combinations and subsequently impose penalties. The application of this strategy is integrated prior to the model's final layer, ensuring smooth integration with most existing PLMs (T5, GPT, LLaMA) without necessitating further adjustments. Implementing our strategy, we observe a significant decline in non-original sequences comprised of more than three words in the academic AASC dataset and the story-based ROCStories dataset. 

\end{abstract}
\section{Introduction}

Pre-trained language models (PLMs) have gained widespread recognition for their unparalleled performance in numerous downstream natural language processing (NLP) tasks \cite{clinchant-etal-2019-use,li-etal-2022-ode,fang-etal-2023-improving,fang-etal-2023-transgec,fang2023chatgpt,zhang2023multitask,pang2024anchor}, especially in text generation \cite{hua-wang-2020-pair,zhang-etal-2020-pointer,guan-etal-2021-long,wang2024what}. 
With the emergence of advanced PLMs, there is an intensifying debate over the distinctiveness of texts produced by these models as opposed to those written by humans, a sentiment highlighted by \citep{mccoy-etal-2023-much}. 
Within this landscape, generative PLMs primarily fall into two architectural paradigms. On the one hand, BART \cite{lewis-etal-2020-bart}, T5 \cite{raffel2020exploring}, and other such pre-trained language models \cite{song2019mass} serve as quintessential representatives of the encoder-decoder approach. 
On the other hand, OpenAI's GPT (Generative pre-trained Transformer) series \cite{radford2019language, NEURIPS2020_1457c0d6,ouyang2022training} and Meta AI's LLaMA series pre-trained language models \cite{touvron2023llama,touvron2023llama2}, leveraging a decoder-only design, has carved out a unique niche for itself. 
Both these architectural categories of pre-trained language models have exhibited impressive capabilities in natural language understanding \cite{ebrahimi-etal-2022-americasnli}, 
%natural language inference \cite{zhang-etal-2020-discriminative}, 
and natural language generation \cite{rothe-etal-2021-simple,jiao2023chatgpt}, solidifying their reputation as premier commercial offerings. 
Because of their efficiency and adaptability \cite{zhan2024prefix}, these models have been widely adopted across diverse sectors, including writing, and academic research. 
Nonetheless, the sophistication these models showcase in text generation has ignited concerns regarding academic integrity, prompting many educational establishments to restrict their utilization in scholarly activities.

\begin{figure}
    \centering
    \includegraphics[width=0.94\columnwidth]{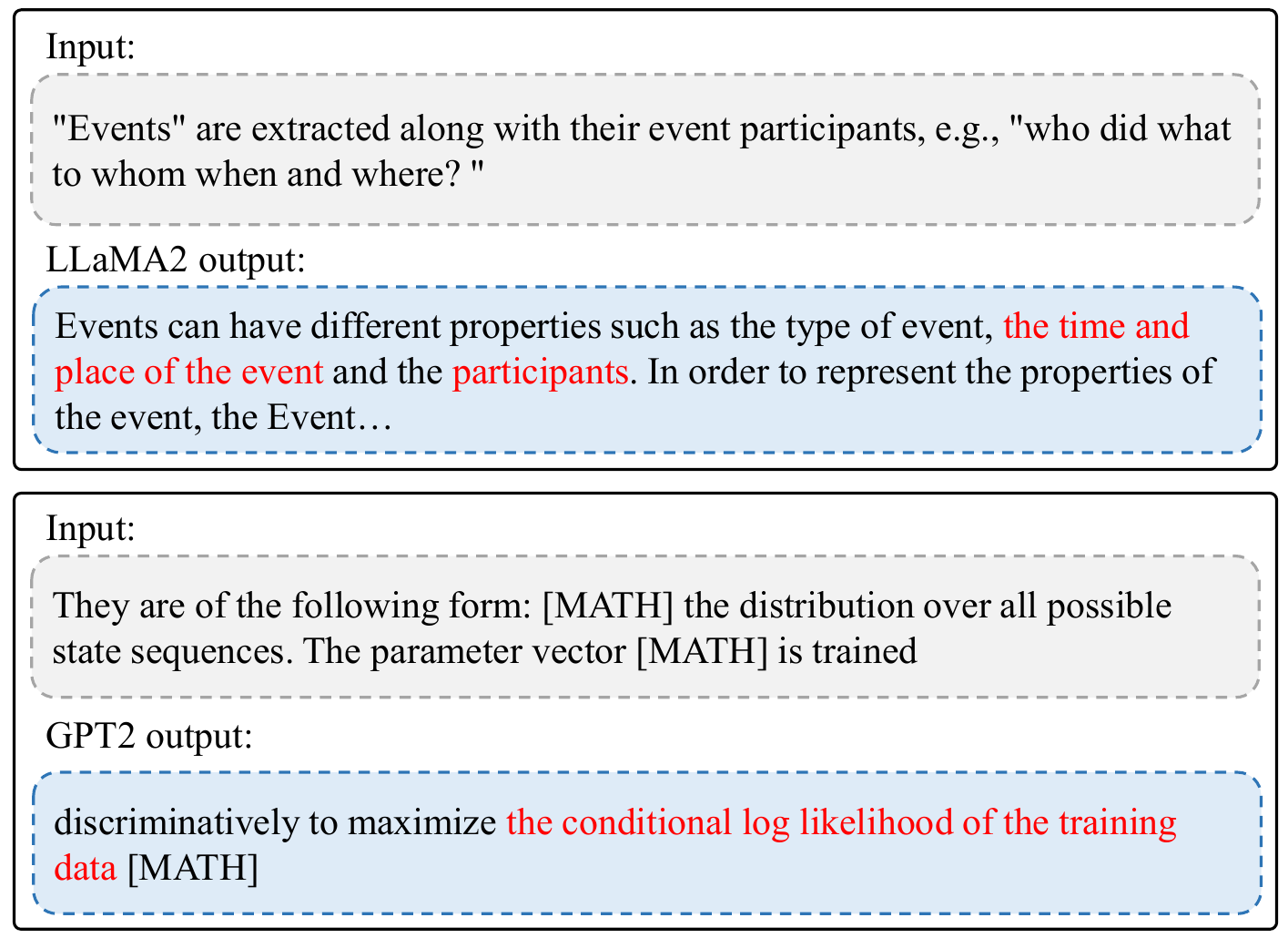}
    \caption{Samples of academic writing generated by the fine-tuned LLaMA2 and GPT2 PLMs using the AASC dataset. Upon plagiarism analysis with Turnitin, both models' outputs showed significant overlaps (in \textcolor{red}{Red}), implying a conspicuous absence of originality. [MATH] indicates a masked math formula.}
    \label{fig:motivation}
\end{figure}

We believe that the aforementioned issues mainly occur because the content generated by language models often lacks originality, as their outputs heavily rely on their training data.
This might lead the models to replicate or mimic the information or patterns they encountered during training. 
In areas where originality is highly valued, such as academic writing or story generation, the outputs from these models can sometimes be seen as plagiaristic.
In fact, earlier research has already highlighted the potential risks of intentional or unintentional leakage of sensitive information within the training sets of language models \cite{Zanella_B_guelin_2020,carlini2021extracting,brown2022does}. This concern persists even during the fine-tuning phases, as evidenced by \citet{mireshghallah-etal-2022-empirical}. To further investigate the potential lack of originality in PLMs outputs, we conduct fine-tuning on two prominent generative PLMs, LLaMA2 and GPT2 pre-trained language models, using an academic paper AASC\footnote{\url{https://github.com/KMCS-NII/AASC}} dataset from the NLP domain. Testing their outputs with Turnitin\footnote{\url{https://www.turnitin.com/}} revealed significant instances where the models reproduced segments from the training set, as illustrated in Figure~\ref{fig:motivation}.

Currently, a considerable portion of research is dedicated to determining whether the outputs from PLMs display plagiarism or retain their originality \cite{ferrero-etal-2017-deep,wahle-etal-2022-large,Lee_2023,wu2024wrote}. 
However, there is a noticeable lack of research focused directly on the innate originality of content generated by these PLMs. To address this void, we introduce a novel generation approach named the "self-plagiarism" (SP) contrastive decoding strategy, aimed at bolstering the inherent originality of text generated by PLMs.
This strategy builds upon the principles delineated by \citet{schick-etal-2021-self} and \citet{chuang2023dola}, yet diverges from their methodologies. Initially, our approach shifts its focus from mitigating model bias at the token level to accentuating originality at the paragraph level. A distinctive feature of our method lies in the strategic emphasis on the topmost layer, enriched with high-level knowledge. Coupled with the extension of generation length, our methodology encompasses a broader spectrum of content uniqueness. This proves particularly advantageous in domains like storytelling and academic writing, where the nuances of high-level knowledge hold significant importance.

Secondly, we achieve an amateur model and a professional model by adjusting the prompts of the language models. For the amateur model, we innovatively introduce three prompts, which originate from the three most common research categories in plagiarized literature: verbatim plagiarism, paraphrase plagiarism, and idea plagiarism. The purpose of these prompts is to guide the model to replicate the training data according to specific plagiarism standards. On the other hand, the professional model uses conventional prompts to encourage the model to generate normal text. On this basis, we subtract the probability distribution of the last layer of the two models and impose penalties on tokens that show a higher probability in a regulatory factor function, ensuring a balance between following the prompts and maintaining originality. Finally, we use our method on various language models, including T5, GPT-2, LLaMA1, and LLaMA2, on the academic dataset AASC and ROCStories dataset. The results show a significant reduction in non-original sequences of more than three words generated by PLMs.

Our primary contributions are as follows:
\begin{itemize}
  \item We reconfirm that even during the fine-tuning phase, pre-trained language models still manifest tendencies of plagiarism. This propensity for un-originality is particularly evident in the domain of academic writing.
  \item We innovatively introduce “self-plagiarism” contrastive decoding strategy by adjusting the prompts of the language models to achieve an amateur model and professional model, and subsequently penalizes the plagiarized tokens. This approach significantly reduces the plagiarism rate and enhances originality.
  \item We showcase the efficacy of our approach in augmenting the originality of the content produced by the models on the widely-used academic writing AASC dataset and ROCStories dataset, providing valuable guidelines for subsequent text generation endeavors.
\end{itemize}

%$h_i = PLM_{\theta}(z_{\lt i})$
\section{Methodology}
\subsection{Definition of Language Modeling}
Pre-trained autoregressive language model $p_{\theta}(y|x)$ is usually built based on the Transformer framework and parametrized by $\theta$. The pre-trained model computes $h_i$ as a function of $z_i$ and the past activations in its left context:

\begin{equation}
    h_i = PLM_\theta(z_{i}),
\end{equation} \\
where $h_i$ is the last layer which is used to compute the distribution for the next token:
\begin{equation}
    p_{\theta}(z_{i+1} | h_{\leq i}) = Softmax(W_{\theta}h_i),
\end{equation}
where $W_{\theta}$ is the pre-trained parameters matrix.

During the finetuning stage, we use the pretrained parameters $\theta$ to initialize the model where $p_{\theta}$ is a trainable language model distribution. The finetuning performs gradient updates on the log-likelihood objective:

\begin{equation}
 \underset{\theta}\arg \max p(y|x;{\theta})  = \sum \log p_{\theta}(z_{i} | h_{< i}),
\end{equation}

\subsection{Plagiarism and Self-plagiarism Prompts}
In the academic domain, plagiarism involves using someone else's work, ideas, or expressions and presenting them as one's own without proper acknowledgment. However, our work focuses on addressing plagiarism in the context of large models—specifically, reducing the straightforward replication of training data by these models. We refer to this as enhancing model's originality.

\begin{figure}
    \centering
   \includegraphics[width=0.9\columnwidth]{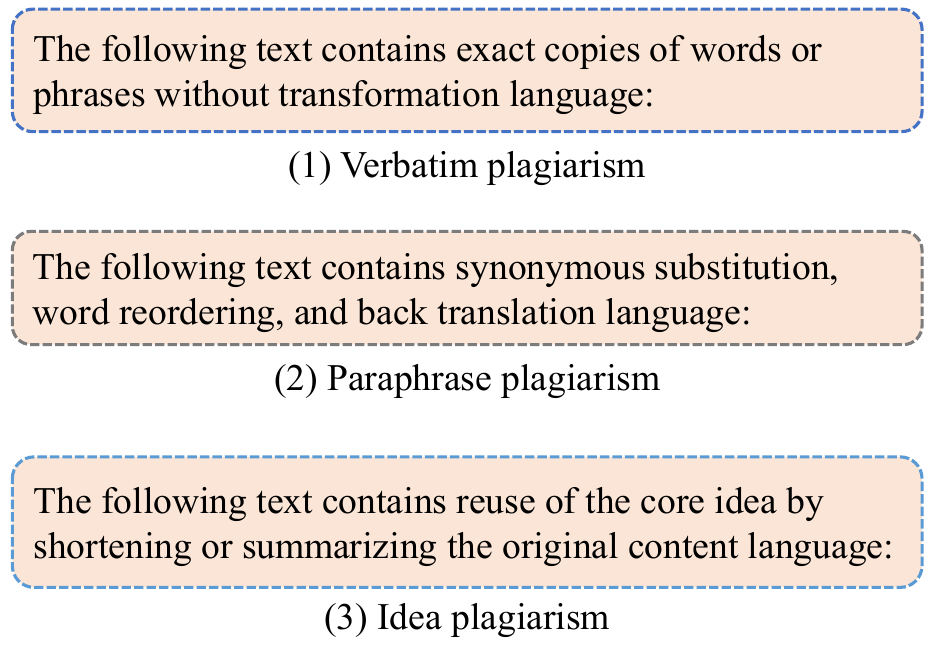}
    \caption{Prompts used for Self-plagiarizing.}
    \label{fig:prompts}
\end{figure}

Inspired by \citet{Lee_2023}, they introduce three most commonly studied categories in plagiarism literature: Verbatim, Paraphrase, and Idea plagiarism. They evaluate both pre-trained and fine-tuned models' plagiarism tendencies in these categories. Their findings suggest language models do not just mimic training samples but can also rephrase or borrow ideas from original texts. Informed by their insights, we craft Self-plagiarizing prompts, tailor to guide the model in replicating training data based on the designated plagiarism criteria. Figure \ref{fig:prompts} illustrates our specific prompt templates.

Formally, let the input text be represented as $X$. Our self-plagiarizing templates encompass text components denoted by $P$, which can manifest in one or more of the ways listed below:

\begin{itemize}
    \item Exact copies of words or phrases without transformation.
    \item Synonymous substitution, word reordering, and back translation.
    \item Reuse of the core idea by shortening or summarizing the original content.
\end{itemize}

\subsection{Amateur LM and Expert LM}
Upon finalizing the design of the plagiarism prompting template, our primary objective is to construct a robustly flawed amateur model, alongside a proficient expert model. In particular, we denote $p(Y |X)$ as the original predictive probability distribution of the input X, and $p(Y|sp(X, P))$ signifies the probability of the subsequent word given input X in self-plagiarizing prompt templates. This self-plagiarizing input urges the language model to manifest plagiarized behavior, laying the foundation for both the amateur and expert models we have developed. The amateur model is established based on $p(Y |X)$, and it emulates self-plagiarizing behavior to formulate a predictive probability distribution $p_{AMA}(Y|sp(X, P))$ for input X. 
Conversely, we provide default system prompts for the expert model, or no prompt at all for models that do not support system prompts. The expert model draws insights from the original predictive probability distribution $p_{EXP}(Y |X)$ to deliver more precise predictions on input X. 

\begin{figure*}[!ht]
    \centering
    \includegraphics[width=1.92\columnwidth]{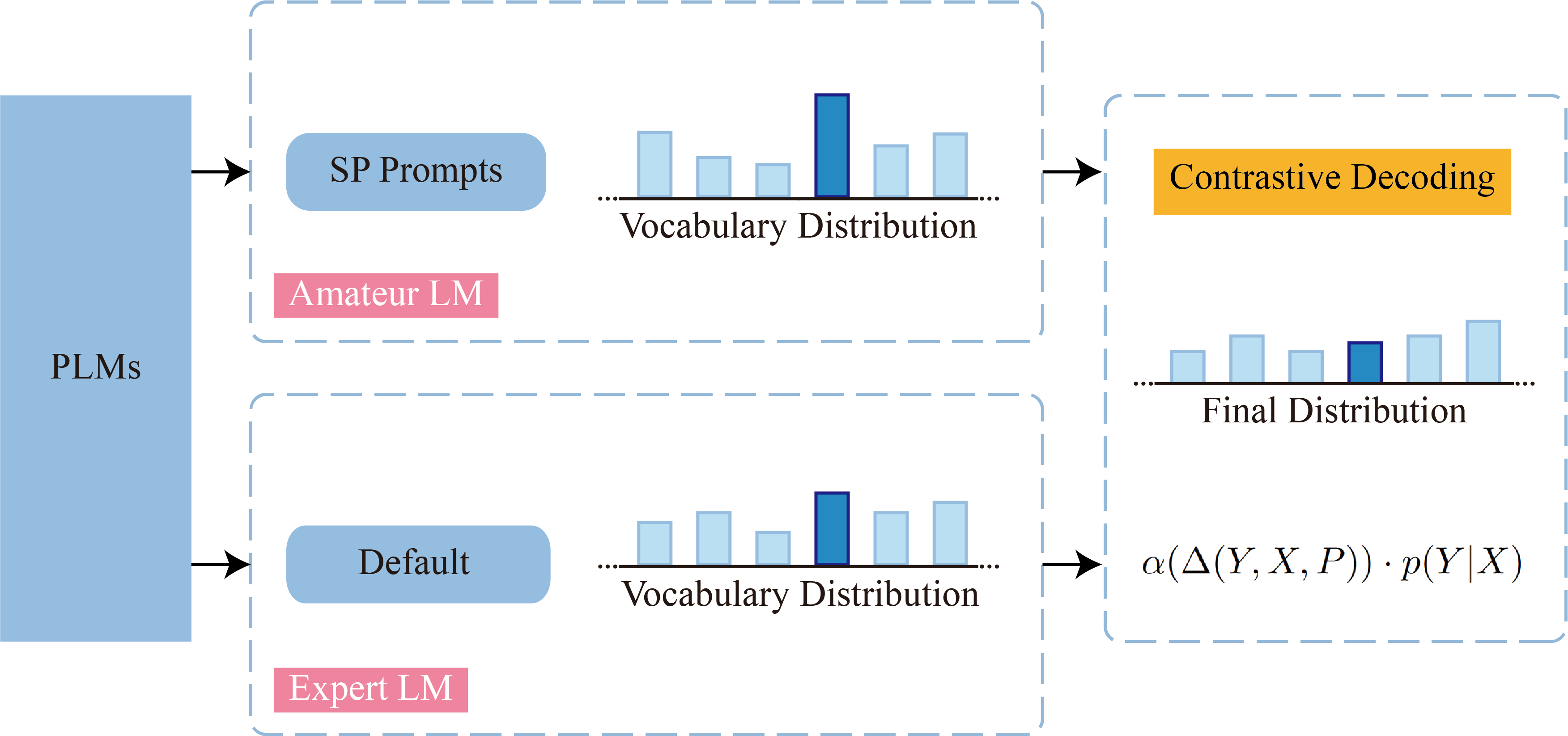}
    \caption{PLMs are prompted to function as both an expert LM and an amateur LM by utilizing SP prompts and default output. The optimized prediction probability is then obtained through contrastive decoding.}
    \label{fig:framework}
\end{figure*}

\subsection{Contrastive Decoding}
Figure \ref{fig:framework} displays the detailed framework of our methodology. The created amateur model tends to produce plagiarized words or fragments more than the expert model. When both the expert model and the amateur model assign a higher probability score to a "repeating token", the expert model is more likely to also assign high scores to other good tokens with low repetitiveness \cite{li-etal-2023-contrastive}. However, the amateur plagiarism model does not behave this way, which means that it is more prone to be influenced by plagiarism prompts.

To make the text generated more original, it is necessary to resist the occurrence of plagiarism. Therefore, we propose a contrastive objective function:

\begin{equation}
    \Delta(Y,X,P)=p_{EXP}(Y|X)-p_{AMA}(Y|sp(X,P)),
\end{equation} \\
In cases of multiple prompts used simultaneously, we keep the largest difference.

\begin{equation}
    \Delta(Y,X,P)=\min_{P}\Delta(Y,X,P),
\end{equation} \\
A plagiarized sentence is more likely to receive a higher probability from $p_{AMA}(Y|sp(X,P))$ than $p_{EXP}(Y|X)$. Hence, for those exact copies of words or phrases, $\Delta(Y,X,P)$ will be lower than zero. For those original expressions, $\Delta(Y,X,P)$ will be greater than zero.

However, it's not necessary to dismiss all results produced by amateur models. To address this, we follow \citet{schick-etal-2021-self} to design a regulatory factor, denoted as $\alpha$, a scale function is used to scale those differences to a number between 0 to 1.

\begin{equation}
\alpha(x)=
        \begin{cases}
            1, \quad &x>0 \\
            e^{\lambda \cdot x}, \quad &otherwise
        \end{cases}
\end{equation} \\
where $\lambda$ is a hyperparameter, $x$ is contrastive objective. For those original expressions ($\Delta(Y,X,P)>0$), their probability will be kept the same, while for those exact copies of words or phrases 
%($\Delta(Y,X,P)<0$), 
, $e^{\lambda \cdot \Delta(Y,X,P)}$ will be lower than 1.

The result will be used to adjust the original predict the probability distribution $p(Y|X)$. 
Finally, model will generate output based on $\tilde{p}(Y|X)$:

\begin{equation}
    \tilde{p}(Y|X)\propto \alpha(\Delta(Y,X,P))\cdot p(Y|X).
\end{equation} 

\section{Experiments}
Considering the vast amount of training data used for pre-trained models and the lack of transparency regarding the specific datasets utilized, it is challenging to detect plagiarize on models that are either not fine-tuned or proprietary. To effectively address the issue, we opt to fine-tune open-source models on publicly available datasets and subsequently assess their plagiarism on these fine-tuned datasets. This strategy allows us to rigorously validate our method and its efficacy in detecting and mitigating plagiarism in PLMs.

\begin{table}[t]
% \linespread{1.15}
\small
\centering
\resizebox{.42\textwidth}{!}{
\begin{tabular}{l|ccc}%一个c表示有一列，格式为居中显示(center)
\toprule
\textbf{Dataset} & \textbf{Train} & \textbf{Eval} & \textbf{Test} \\
\cmidrule(lr){1-4}
 ROCStories &  98,161 & 1572 & 1,871  \\
 AASC & 282,332  & 2812 & 2,874  \\
\bottomrule
\end{tabular}}
\linespread{1}
\caption{Summary of the Experimental Datasets}
\label{tab:datasets}
%\vskip -1em
\end{table}

\subsection{Dataset}
We utilize the ROCStories training dataset \cite{mostafazadeh-etal-2016-corpus} and the ACL Anthology Sentence Corpus (AASC)\footnote{https://github.com/KMCS-NII/AASC} during the fine-tuning phase of the language models.
For the ROCStories dataset, we strictly follow the partition outlined in \cite{mostafazadeh-etal-2016-corpus}, dividing it into training, testing, and development sets.
The AASC dataset constitutes a curated compilation of text excerpts extracted from scientific papers in the field of natural language processing. Drawn from PDF-format papers published within the ACL Anthology between 2000 and 2018, each paper is segmented into individual sentences, categorized according to their respective originating sections. To construct our training set, we chose sections including Abstract, Introduction, Background, Method, Result, and Discussion. For evaluation and testing, a chosen 1\% of sentences were randomly extracted from the corpus.
%, creating distinct evaluation and testing subsets.
A detailed data statistics can be found in Table \ref{tab:datasets}.

\subsection{Model and Training}
With due consideration for performance optimization and the judicious allocation of computational resources, our methodological framework revolves around the utilization of LLaMA-7b series pre-trained language models \cite{touvron2023llama,touvron2023llama2}, GPT-2 large \cite{radford2019language} and T5 large \cite{raffel2020exploring} as baseline models. Our training process leverages the computational power of four A100 or V100 GPUs. Both GPT-2 and T5 models undergo fine-tuning over a span of five epochs, employing two distinct datasets. Because of limitation of computing resource, LLaMA series are fine-tuned over 3 epochs with Alpaca-LoRA.\footnote{https://github.com/tloen/alpaca-lora} The foundational pre-trained models and training scripts are sourced from the Huggingface repository.\footnote{https://github.com/huggingface/transformers} 
Appendix \ref{appendix:hyper-parameters} presents a detailed account of the hyper-parameters used in the training process.

\subsection{Evaluation}
We employ three distinct evaluation ways: Generation Originality Test (GOT) \cite{brooks-youssef-2021-got}, Turnitin\footnote{https://www.turnitin.com}, and human evaluation. GOT is an n-gram automated test for assessing originality. It constructs an original set by extracting unique n-gram fragments from the training set and subsequently examines whether the fragments generated in the test output are contained within this original set (see Appendix \ref{appendix:GOT}). We apply GOT to all models on both datasets.
To further evaluate if our approach effectively mitigates plagiarism in academic writing and enhances the model's originality, we utilize Turnitin, a widely used academic plagiarism detection software, specifically on the AASC academic dataset.
For human evaluation, we enlist the feedback of two volunteers to assess the impact of our method on the coherence (the logical connection and content association between sentences) and fluency (grammar error, naturalness, and writing style) of the model's output. Appendix \ref{appendix:human evaluation} presents a detailed definition of coherence and fluency, as well as evaluation template and examples.

\begin{figure*}[!ht]
    \centering
    \includegraphics[width=1.92\columnwidth]{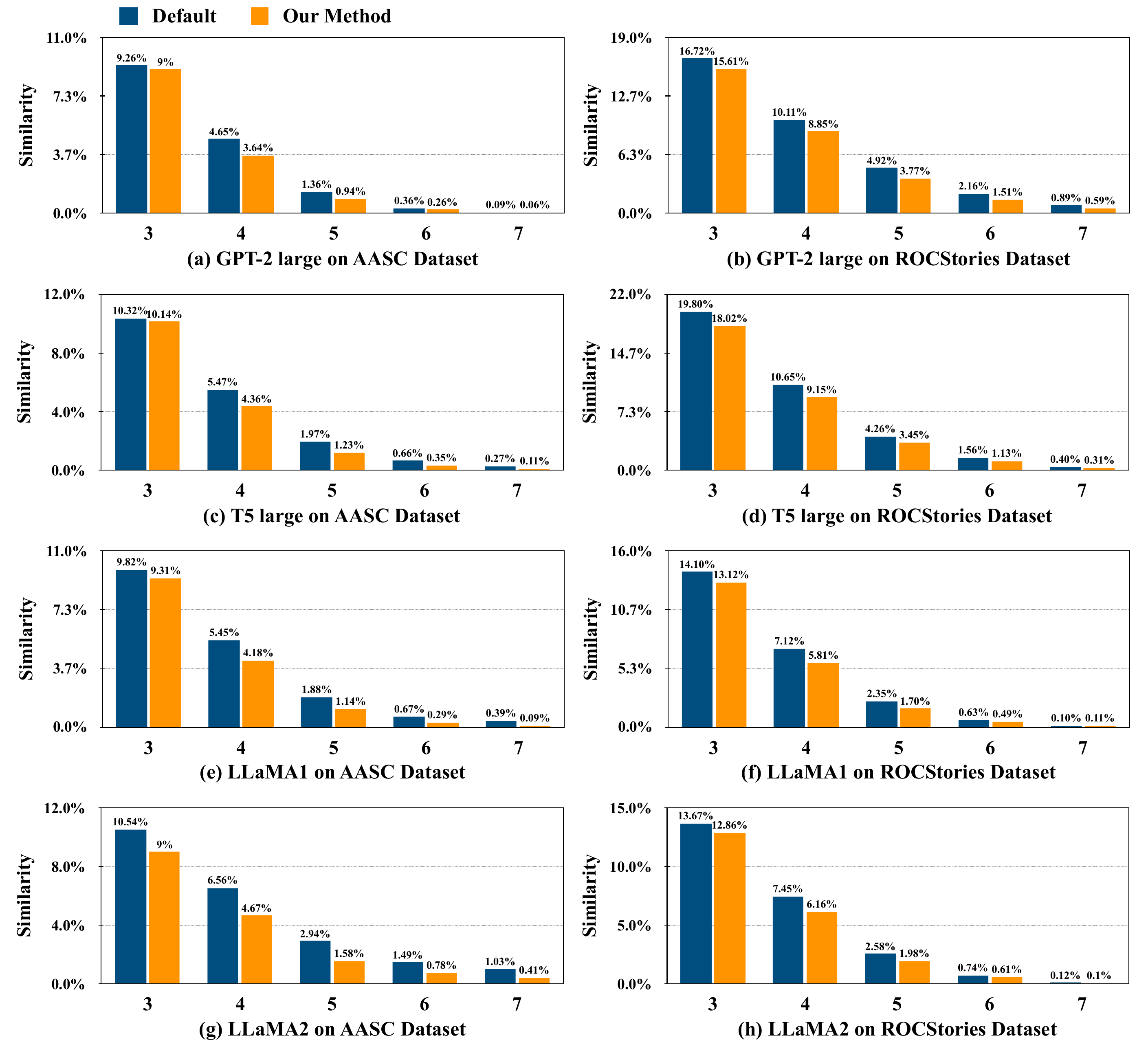}
    \caption{Evaluation results of the fine-tuned GPT-2 large, T5 large, LLaMA1 and LLaMA2 PLMs on the ROCStories and AASC datasets using GOT Metric. Our proposed method exhibits reduced plagiarism across various fragments lengths compared to the default, highlighting its enhanced originality.}
    \label{fig:result}
\end{figure*}

\section{Results}
\subsection{Evaluation on GOT}
% Figures \ref{fig:result}, \ref{fig:result-llama-aasc} and \ref{fig:result-llama-roc} 
Figures \ref{fig:result} illustrate the comparative analysis of output dissimilarity between the four pivotal models, evaluated using the GOT metric. Appendix \ref{appendix:Example outputs} presents illustrative instances of input and output.
In terms of overarching trends, a discernible pattern emerges: the resemblance between model-generated texts diminishes in opposite proportion to the length of the identified segments. The occurrence of text displaying similarity, spanning more than seven consecutive words, becomes exceedingly rare. Moreover, the similarity reduction due to the SP contrastive decoding strategy decreases with increasing segment length. All of the four models show a lower similarity rate on AASC than on ROCStories. We think the reason lies on ROCStories has a more common used vocabulary, while the inherent limitations of the SP contrastive decoding strategy inadvertently constrain the range of potential candidates for model predictions, prompting a selection bias towards words that are more prone to detection by the GOT algorithm.

Within the context of the dataset evaluated through ROCStories analysis, LLaMA series show an overall lower similarity, indicates a better originality of large model. GPT-2 large consistently exhibits a reduction of over one percent in similarity across all segment lengths up to five words. In contrast, T5 large deviates from this trajectory, exhibiting less similarity decreases when fragment length is higher than five, but maitain a lower similarity rate at the same time, which means T5 takes an advantage on similarity when generating a long sentence, also makes a discount on the performance of SP contrastive decoding strategy.

\subsection{Evaluation by Turnitin Check on AASC test set}
We collate the output texts generated by LLaMA2, GPT-2 and T5 on the AASC dataset. Subsequently, these generated texts are systematically uploaded onto the Turnitin platform to undergo rigorous originality assessments. Finally, we analyze the outcomes of these Turnitin evaluations. Table \ref{tab:turnitin} showcases the assessment outcomes rendered by Turnitin. For all of the three fine-tuned pre-trained models, our proposed method yields a at least 3\% decrease in similarity rates. This empirical validation through Turnitin lends robust support to the practical efficacy of our proposed SP contrastive decoding strategy, confirming its effectiveness within real-world applications.

\begin{table}[t]
\centering
\resizebox{0.98\columnwidth}{!}{
\begin{tabular}{@{}cccc@{}}
\toprule
%\multicolumn{3}{c}{Turnitin Similarity Check}  \\ \midrule
\multicolumn{1}{c|}{Model} & \multicolumn{1}{c|}{LLaMA2} & \multicolumn{1}{c|}{GPT2 large} & T5 large \\ \midrule
\multicolumn{1}{c|}{Default} & 12\% & 5\%  & 6\%  \\
\multicolumn{1}{c|}{Our Method} & 4\% & 2\% & 3\%  \\ \bottomrule
\end{tabular}
}
\caption{Turnitin Similarity Check on the AASC dataset. Our method consistently achieves a minimum 3\% reduction in the similarity of the model's output text.}
\label{tab:turnitin}
%\vskip -1em
\end{table}

\subsection{Human Evaluation}
We average the scores of the two volunteers for the model evaluation. 
The results, outlined in Table \ref{tab:human_evaluation}, demonstrate that, in the majority of cases, our method yields comparable results in coherence and fluency. Notably, it even exhibits higher coherence on the AASC dataset. We attribute this to the robustness of pre-trained language models (PLMs), which can generate coherent and fluent text even when certain tokens are penalized. This suggests that our method has minimal negative effects on the coherence and fluency of the model's output, successfully preserving the natural flow and understandability of the generated content.

\begin{table}
\resizebox{0.98\columnwidth}{!}{
\begin{tabular}{l|cccc}
\toprule
% \hline
                       & \multicolumn{2}{c|}{\textbf{Coherence}} & \multicolumn{2}{c}{\textbf{Fluency}} \\ \midrule %\hline
                       & default    & \multicolumn{1}{c|}{SPCD}   & default           & SPCD              \\ \midrule %\hline
A\_T5        & 83.5\%     & 88.5\%                    & 87.3\%            & 87.2\%           \\
A\_LLaMA2           & 86.5\%     & 88.0\%                     & 82.0\%            & 83.0\%           \\
R\_GPT2 & 76.8\%     & 76.8\%                     & 74.5\%            & 75.3\%           \\
R\_LLaMA2     & 92.3\%     & 87.5\%                     & 87.5\%            & 85.3\%           \\ \bottomrule %\hline
\end{tabular}
}
\caption{Human evaluation of coherence and fluency on LLaMA2, GPT2-Large and T5-Large. The notation ``A\_model'' refers to the fine-tuned model using the AASC dataset, while ``R\_model'' denotes the fine-tuned model using the ROCStories dataset, SPCD means SP contrastive decoding strategy.}
\label{tab:human_evaluation}
\end{table}

% \section{Analysis}
\section{Ablation study}
\subsection{Impact of Self-plagiarism Prompts}
\begin{figure}[t]
    \centering
    \includegraphics[width=0.98\columnwidth]{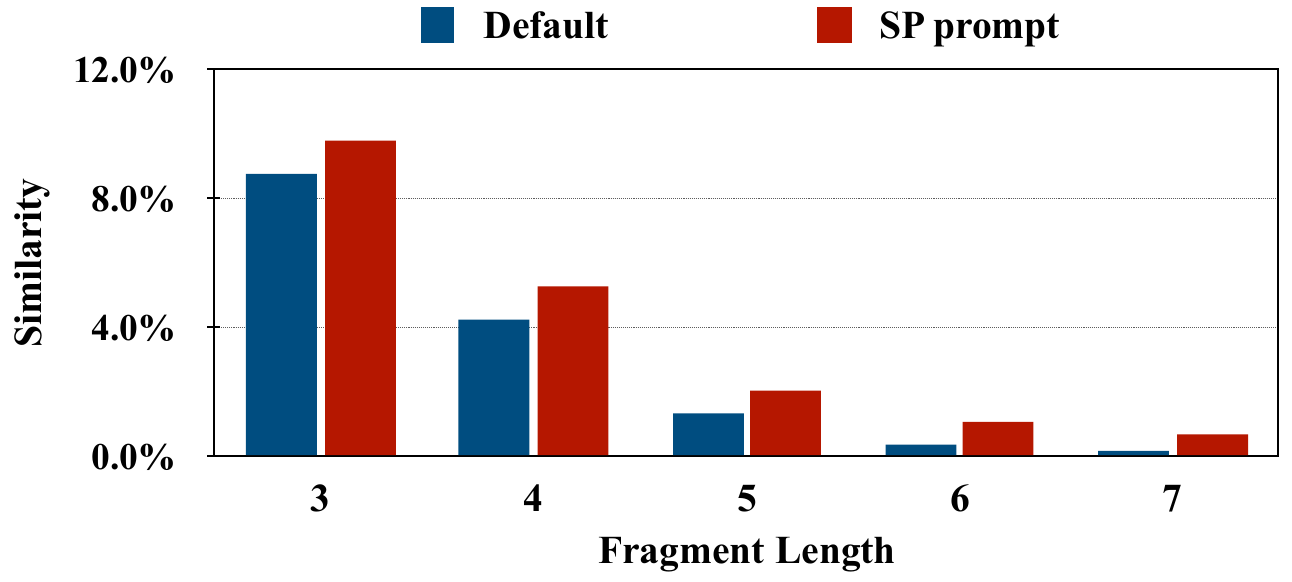}
    \caption{Impact of Self-plagiarism Prompts. When adding only self-plagiarism prompt, the similarity score consistently increase.} 
    % \textbf{SP} indicates outputs with self-plagiarism prompt.}
    \label{fig:Ablation Self-Plagiarism}
\end{figure}

To independently verify the impact of self-plagiarizing prompts, we design ablation experiments using LLaMA2 with both default system prompts and self-plagiarizing prompts, specifically: 
\begin{itemize}
    \item The following text contains exact copies of words or phrases without transformation of language:
\end{itemize}
We then compare the similarity of the output results. The findings, depicted in Figure \ref{fig:Ablation Self-Plagiarism}, indicate that the similarity of the outputs generated with self-plagiarizing prompts is higher across all fragment lengths compared to those generated with default prompts. This demonstrates the effectiveness of the contrastive decoding strategy.

\subsection{Impact of Varying Prompt Numbers}
\begin{figure}[t]
    \centering
    \includegraphics[width=0.98\columnwidth]{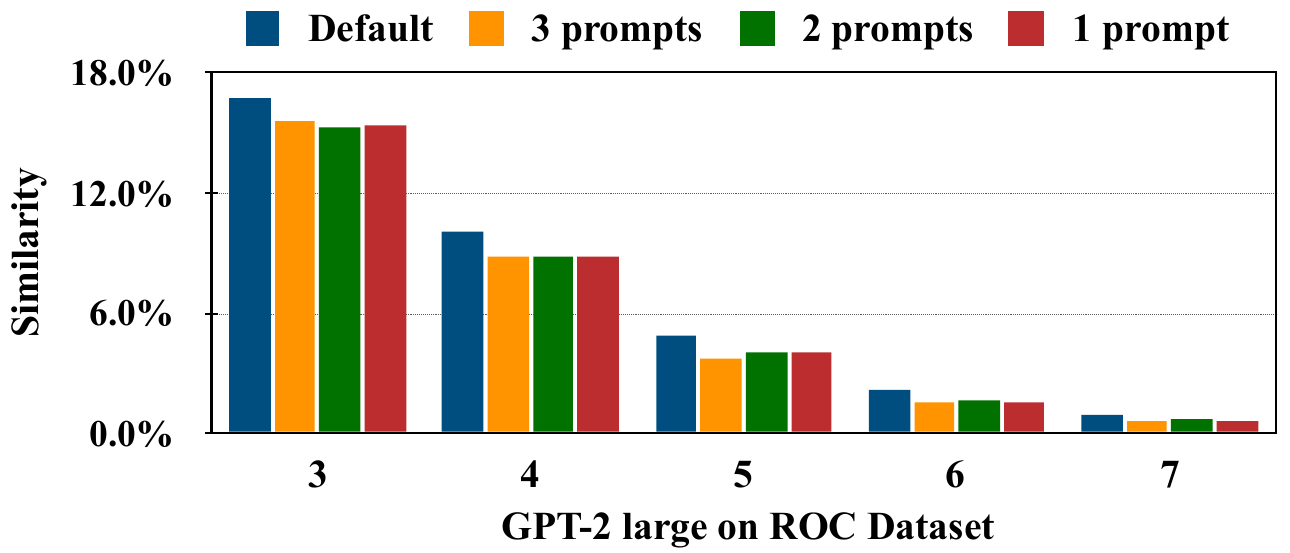}
    \caption{Impact of Varying Prompt Numbers. When adding three prompts, the similarity score reaches its lowest point.}
    \label{fig:Ablation}
\end{figure}

To comprehensively comprehend the influence of prompt quantity on the SP contrastive decoding strategy's efficacy, we conduct a meticulous evaluation using GPT-2 large PLM on the ROCStories dataset, modulating the number of prompts as the independent variable. As depicted in Figure \ref{fig:Ablation}, our findings reveal a consistent trend: the incorporation of the SP contrastive decoding strategy invariably leads to a reduction in the similarity of the generated texts across all test instances. This demonstrates that our approach enhances the originality of content produced by PLMs.

Upon augmenting the prompts to three, a slight augmentation in performance arrives. This enhancement emerges when the fragment length exceeds three words. This empirical observation underscores the pivotal role that sufficiently comprehensive prompts play in adeptly guiding the model's generation process. This nexus between prompt comprehensiveness and algorithmic efficacy underscores that the SP contrastive decoding strategy can achieve heightened performance levels with a more elaborate array of prompts. This augmentation manifests as a reduction in similarity across an array of test cases, reinforcing the integral role of robust and all-encompassing prompts in enhancing the overall efficacy of the SP contrastive decoding strategy.

\begin{figure}[t]
    \centering
    \includegraphics[width=0.98\columnwidth]{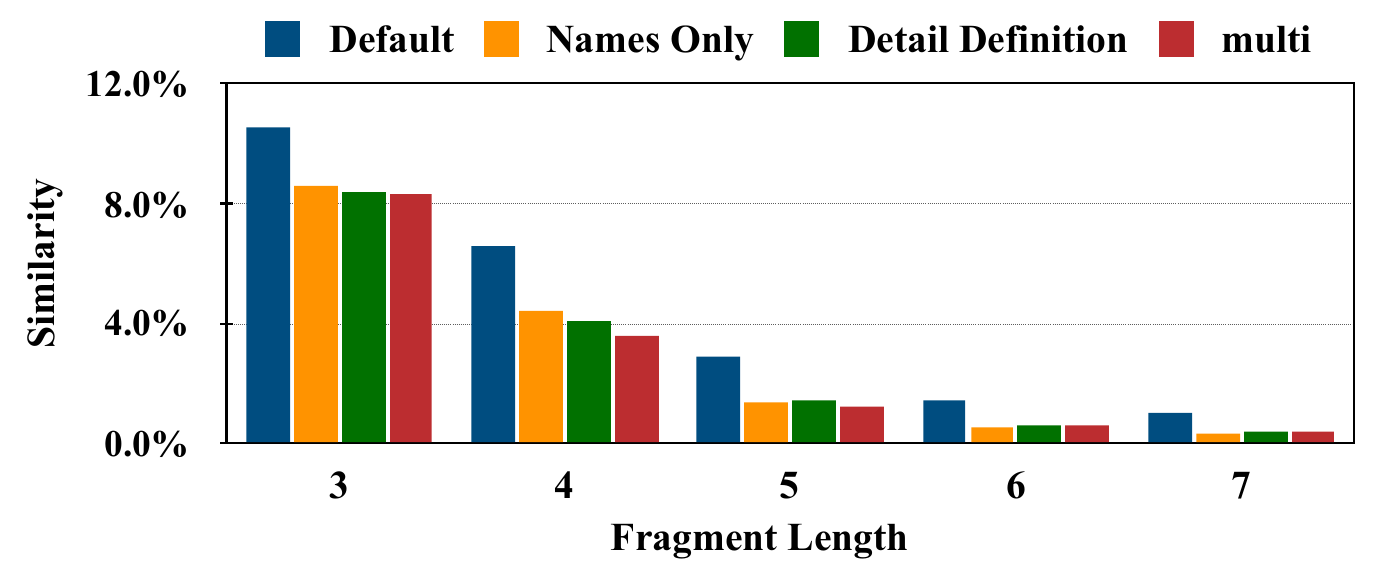}
    \caption{Impact of Prompt Template. A more specific prompt, along with a higher quantity of prompts, contributes to a more significant improvement in performance.}
    \label{fig:2-template}
\end{figure}

\subsection{Impact of Prompt Template}
We evaluate the performance of LLaMA2 on the AASC dataset using various prompts, as depicted in Figure \ref{fig:2-template}. In the "Name Only" template, we streamline the detailed definitions of the three types of plagiarism to include only their names, while the "Detail Definition" template retains the same format as in the previous experiment. Importantly, the multi-prompt condition, which includes both templates, outperformes other control groups. These findings highlight that a more specific prompt and a higher quantity of prompts contribute to a more substantial improvement in performance.

\section{Related Work}
\subsection{Plagiarism and Memorization in PLMs}
Numerous studies have consistently shown that pre-trained language models have a tendency to memorize and plagiarize content from their training data. The study by \citet{brown2022does} underscores the potential risk of intentional or unintentional disclosure of sensitive information from a model's training set. Pioneering research, such as that by \citet{Zanella_B_guelin_2020} and \citet{carlini2021extracting}, has revealed the extensive ability of large-scale models to internalize training samples during pre-training, making them especially susceptible to membership inference and data extraction attacks. Importantly, a significant portion of the training datasets used for language model training are culled from the Internet, often without obtaining clear consent from the original content creators \citet{brown2022does}. \citet{Lee_2023} evidence that PLMs do reproduce content from training samples, encompassing all three classifications of plagiarism. More recently, \citet{mccoy-etal-2023-much} explore the novelty of machine-produced texts and conclude that neural language models have the knack for weaving familiar components into fresh content, rather than merely echoing training samples. Collectively, these studies illuminate the undeniable fact that the outputs of PLMs do draw heavily from their training samples. This sheds light on the deep reliance of pre-trained language models on their training data, underscoring their limitations in terms of innovation and originality.

\subsection{Plagiarism Detection for LMs}
Plagiarism detection in language models refers to determining whether the model's output replicates content from its training data. To identify if machine-generated texts directly ``plagiarize" from its training set, researchers have developed a multitude of detection techniques. For instance, \citet{bensalem-etal-2014-intrinsic} introduce an innovative, language-agnostic plagiarism detection approach, dubbed the n-gram class method, which relies on a novel text representation. \citet{Kppers2012ASA} gauge the Dice coefficient for 250-character blocks between passage pairs, while \citet{Shrestha2013UsingAV} employ n-grams to compute the Jaccard similarity. Relying on Convolutional Neural Networks, \citet{Agarwal_2018} extract regional information from n-grams and use Recurrent Neural Networks to grasp long-term dependencies. \citet{Alzahrani2015ArabicPD} identify candidate documents by searching for exact duplicated sequences and analyze the similarity of overlapping 8-grams. In the context of generation tasks lacking standard automatic measures, \citet{brooks-youssef-2021-got} propose an automated originality testing method. Recently, \citet{Lee_2023} embrace a combination of traditional similarity metrics and cutting-edge models, aiming to enhance the efficacy of plagiarism detection. While these studies focus on pinpointing models' plagiaristic behaviors, the question of how to amplify the originality of the generated content has remained unexplored. 

\subsection{Contrastive Decoding}
\citet{li-etal-2023-contrastive} use an amateur language model (LM) to aid an expert LM in generating coherent text. The work of \citet{obrien2023contrastive} additionally supports the efficacy of contrastive decoding in text generation.
In our approach, we integrate their concept into downstream tasks. 
We use prompts to encourage model plagiarism. This allows us to utilize a single PLM, comparing its predictions when conditioned by opposing prompts, thereby discerning and refining the model's responses. 
In a related vein, \citet{chuang2023dola} propose a contrastive decoding approach, emphasizing the disparity in logits between a higher layer and a lower layer to derive the output probability over the next word. 
Additionally, \citet{dai-etal-2022-knowledge} discover the presence of ``knowledge neurons'' distributed predominantly in the topmost layers of pre-trained BERT models. 
Building upon these insights, our work adopts a similar idea, but just leveraging knowledge from the topmost layer to penalize plagiarized tokens. This will contribute to the improving of originality of model's generation.

\section*{Conclusion}
In this study, we emphasize once again the susceptibility of fine-tuned PLMs to display plagiarism tendencies and replicate content from their training sets. This observation applies to both encoder-decoder and decoder-only architectures.
These findings highlight the potential risks linked to the utilization of PLMs in sensitive domains like academic writing and storytelling.
To tackle the problem of model plagiarism and boost its originality, we introduce an approach involving the introduction of three plagiarism prompts.
These prompts guide the model to initially replicate training data in accordance with plagiarism prompts, effectively functioning as an amateur model. 
Simultaneously, penalties are applied to tokens that show increased probabilities in a designated function, thereby enhancing the overall originality of the generated text.
Implementing the proposed approach results in a significant reduction in the generation of non-original sequences, particularly those with over three-word fragments, across both the AASC and ROCStories datasets using PLMs.
The alignment of these outcomes offers a robust evaluation of our proposed SP contrastive decoding strategy's ability to ensure the originality of generated text. 
Furthermore, human evaluation suggests that our method has minimal adverse effects on the coherence and fluency of the output from large pre-trained models.
This comprehensive validation highlights the strength and practical utility of our proposed approach in addressing plagiarism risks in NLP tasks.

\section*{Limitations}
Due to computational limitations, we can not test SP contrastive decoding strategy on larger pre-trained models like GPT-3 \cite{NEURIPS2020_1457c0d6}. This will be a focus of our future work.
Given the opacity of the training data and the high complexity of the GOT algorithm, we opt not to validate our method on base models without fine-tuning. This is an area we plan to address and improve upon in future work.
The primary constraint inherent to our approach lies in its theoretical inability to entirely eradicate plagiarism. Furthermore, given its prompt-based nature, the efficacy of the SP contrastive decoding strategy is contingent upon the model's inherent comprehension capacity. This dependence implies that its performance might be less optimal when applied to models outside the realm of PLMs. 
Another potential concern with the SP contrastive decoding strategy is its inherent need to prompt the model to generate predictions contrary to the intended outcome. This complicates the algorithm's use for purposes other than counter-plagiarism unless a consistently opposing directive is employed.

\section*{Ethics Statement}
This research endeavor rigorously adheres to ethical guidelines and principles. 
We exclusively employ publicly accessible models, datasets, and tools, thereby precluding any involvement or collection of sensitive or private data.  
All datasets and model parameters are obtained strictly for research purposes from public repositories. 
Central to this study is the objective to chart a course for effectively guiding the behavior of models, aiming to enhance their capacity to assist users ethically. 
Our ambition is to foster the ethical advancement of language models, championing their deployment in ways that respect ethical standards and promote societal well-being.

\section*{Acknowledgements}
This work was supported in part by the Science and Technology Development Fund, Macau SAR (Grant Nos. FDCT/0070/2022/AMJ, FDCT/060/2022/AFJ), Ministry of Science and Technology of China (Grant No. 2022YFE0204900), National Natural Science Foundation of China (Grant No. 62261160648), the Multi-year Research Grant from the University of Macau (Grant No. MYRG-GRG2023-00006-FST-UMDF), and Tencent AI Lab Rhino-Bird Gift Fund (Grant No. EF2023-00151-FST). This work was performed in part at SICC which is supported by SKL-IOTSC, and HPCC supported by ICTO of the University of Macau.

%\clearpage
\bibliography{custom,anthology}
% \bibliography{anthology,custom}
% \bibliographystyle{acl_natbib}

\clearpage
\appendix
\label{sec:appendix}

\section{Hyper-parameters for training}
\label{appendix:hyper-parameters}
As shown in table \ref{tab:Hyper-parameters for training}, in the fine-tuning process of both LLaMA1 and LLaMA2 using both the AASC and ROCStories datasets, specific hyperparameters were employed. These include a learning rate of 1e-4, a batch size of 4, a LORA-R of 8, a LORA-Alpha of 16, and a LORA-Dropout of 0.05.
For the fine-tuning of GPT-2 with the AASC dataset, a learning rate of 5e-4 is adopted. The block size is configured at 128, and a batch size of 16 per device is instituted, further compounded by a gradient accumulation step count of 2. Conversely, for the fine-tuning of GPT-2 using the ROCStories dataset, default learning rate settings are maintained. The block size is established at 60, with a batch size of 128 per device and a gradient accumulation step count of 2.
As for the T5 model, %the source and target languages are designated as ``up" and ``down" respectively. The source prefix is ``translate up to down: ". 
the batch size is set to 24 per device. During the fine-tuning phase on the AASC dataset, a learning rate of 0.001 is employed, coupled with a gradient accumulation step count of 4. Correspondingly, for the ROCStories dataset, the learning rate is maintained at 1e-4, and the gradient accumulation step count is retained at 2.

\section{GOT Algorithm}
\label{appendix:GOT}
Table \ref{tab:GOT-algorithm-2} provides the pseudo code of the Generation Originality Test (GOT) algorithm. We utilized this as the basis for implementing the Python script of the GOT algorithm for conducting similarity testing.

\begin{table}[!ht]
\resizebox{\columnwidth}{!}{
\begin{tabular}{@{}lp{\columnwidth}@{}}
\toprule
\multicolumn{2}{p{\columnwidth}}{\textbf{GOT Algorithm}: Detect similar fragments}  \\ 
\midrule
\multicolumn{2}{p{\columnwidth}}{\textbf{Input}: Input Sentences $X$, List of the sentences $S$, Original set $O$ }      \\
\multicolumn{2}{p{\columnwidth}}{\textbf{Result}: Similar fragments $R$ }      \\

    % 1   & count = empty list                                       \\
    1   &  Initialize $R$ as empty list                                      \\
    2   & \textbf{foreach} sentence \textbf{in} $X$ \textbf{do}                \\
    3   & \quad Get length of sentence as $sl$                                 \\
        & \quad // Define a sliding window of length $wl$     \\
    4   & \quad \textbf{for} $wl$ = 2 to $sl$ \textbf{do}       \\
        & \quad \quad // Move the window, and extract fragments \\
    5   & \quad \quad \textbf{for} $i$ = 0 to $sl$ - $wl$ +1 \textbf{do}             \\
    6   & \quad \quad \quad Extract fragment from sentence                  \\
    % 8   & \quad \quad \quad count{[}window{]} += 1                                   \\
    7   & \quad \quad \quad \textbf{if} fragment \textbf{in} $O$ \textbf{then}         \\
    8   & \quad \quad \quad \quad Add fragment to $R$ \\ 
    9   & \textbf{return} $R$ \\
\bottomrule
\end{tabular}
}
\caption{GOT Algorithm.}
\label{tab:GOT-algorithm-2}
\end{table}

\begin{table*}[htbp]
\centering
\begin{tabular}{l|ccccc}
\toprule %\hline
Config.      & LLaMA1 \& LLaMA2    & \multicolumn{2}{c}{GPT-2 Large} & \multicolumn{2}{c}{T5 Large} \\ \midrule %\hline
Dataset      & AASC \& ROCStories & AASC          & ROCStories      & AASC         & ROCStories    \\
GPU          & A100 * 4           & V100 * 4      & V100 * 4        & V100 * 4     & V100 * 4      \\
Learning Rate           & 1e-4               & 5e-4          & 1e-5            & 1e-3         & 1e-4          \\
Batch Size   & 4                & 16            & 128             & 24           & 24            \\
Epochs       & 3                  & 5             & 5               & 5            & 5             \\
Block Size   & -                   & 128           & 60              &  -            &  -             \\
Gradient Accu.         &  -                  & 2             & 2               & 4            & 2             \\
LORA-R       & 8                  &  -             &   -              &  -            &  -             \\
LORA-Alpha   & 16                 &  -             &  -               &  -            &  -             \\
LORA-Dropout & 0.05               &  -            & -                &    -          &  -             \\ \bottomrule %\hline
\end{tabular}
\caption{Hyper-parameters for training LLaMA1, LLaMA2, GPT-2 Large and T5 Large models. Gradient Accu. means gradient accumulation steps.   - means not applicable or default setting.} %LR refers to learning rate, GA means gradient accumulation step count. A blank means not applicable or default setting.}
\label{tab:Hyper-parameters for training}
\end{table*}

\section{Human Evaluation}
\label{appendix:human evaluation}
\subsection{Evaluation Metric}
In this section, we provide definitions for coherence and fluency, breaking them down into specific components:\\\\
Coherence:
\begin{itemize}
    \item Content Association: Indicates whether the output sentence is related to the content of the source sentence.
    \item Logical Coherence: Rates the natural degree of logical connection between the output sentence and the source sentence.
\end{itemize}
Fluency:
\begin{itemize}
    \item Grammar Errors: Indicates whether the output sentence contains grammar errors.
    \item Naturalness and Vividness: Assesses whether the language of the output sentence is natural and vivid, as opposed to being stiff or verbose.
    \item Writing Style: Examines whether the output sentence aligns with the style of academic papers or narrative storytelling.
\end{itemize} 
\subsection{Template of Human Evaluation}
\label{appendix:Template of Human Evaluation}
\begin{figure*}[!ht]
    \centering
    \includegraphics[width=1.9\columnwidth]{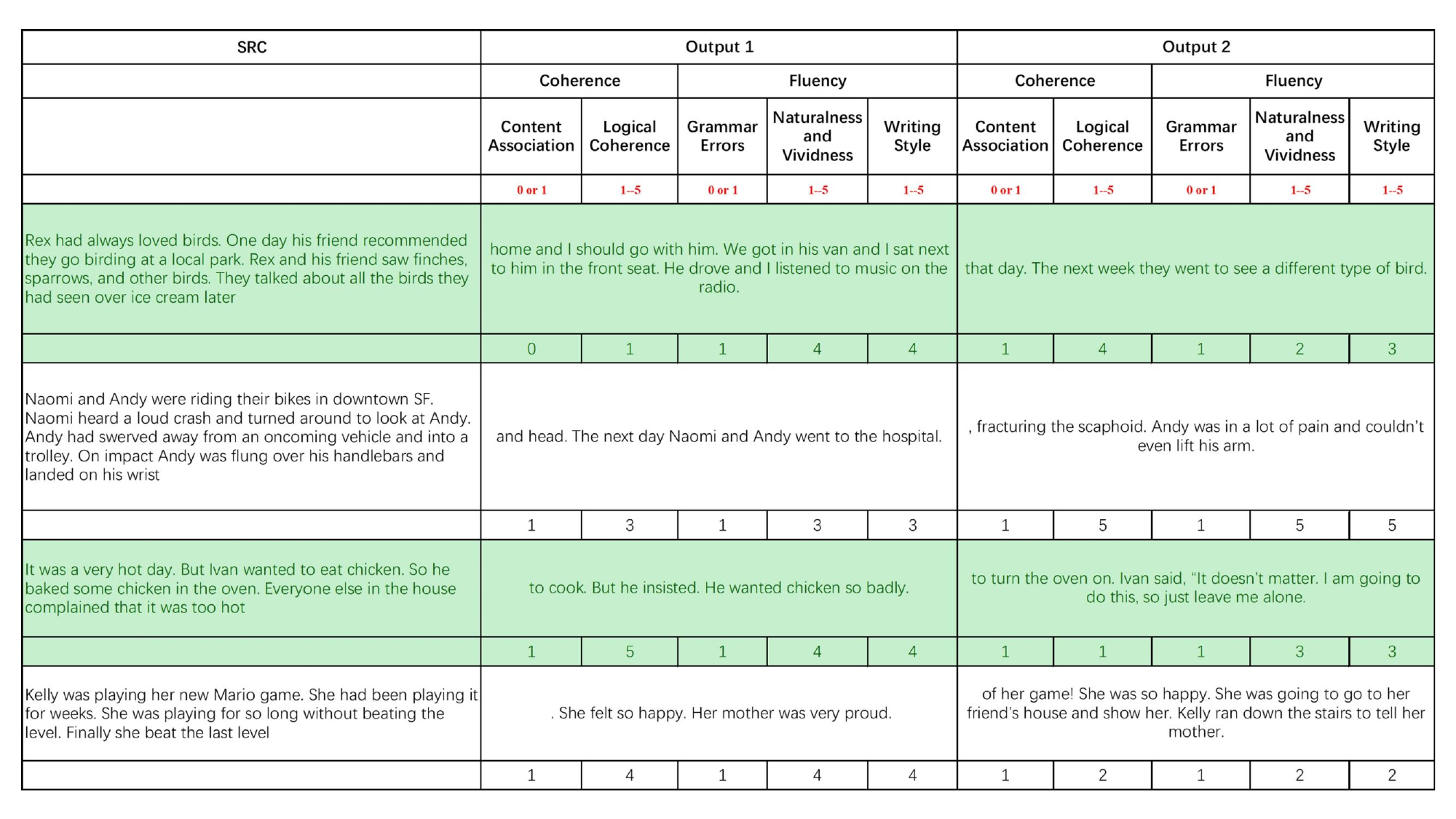}
    \caption{Template and examples of human evaluation.}
    \label{fig:Human evaluation}
\end{figure*}

Figure \ref{fig:Human evaluation} illustrates the template utilized for manual evaluation alongside four concrete examples. It is pertinent to note that due to the recruitment of two volunteer assessors who are native Chinese-speaking doctoral candidates, certain rule explanations are presented in Chinese. The test sentences were randomly sampled from model outputs. Additionally, the identities of the three models have been concealed to uphold the integrity and fairness of the evaluation process. This methodology ensures an unbiased assessment of model performance, safeguarding against potential predispositions or biases associated with prior knowledge of specific model identities.

% \begin{figure*}
%     \centering
%     \includegraphics[height=13.5cm, angle=270]{figures/Human-Evaluation-Examples-en.jpg}
%     \caption{Template and examples of human evaluation.}
%     \label{fig:Human evaluation}
% \end{figure*}

\section{Example outputs}
\label{appendix:Example outputs}
Table \ref{tab:case study ROC} and \ref{tab:case study AASC} elucidates a collection of exemplars encompassing input texts and corresponding outputs, both with and without the integration of the SP contrastive decoding strategy, derived from both fine-tuned LLaMA2, GPT-2 large and T5 large pre-trained models across the two distinct datasets (ROCStories and AASC datasets). Evidently, all instances of SP contrastive decoding strategy generated outputs aptly maintain logical coherence and textual fluency while effectively circumventing plagiarism. This substantiates the inherent compatibility of the SP contrastive decoding strategy with models underpinning disparate architectures, as well as its efficacy within divergent text generation domains. This robustness further underscores the algorithm's commendable generalization potential.
However, it is imperative to acknowledge a potential drawback: the proposed SP contrastive decoding strategy's direct influence on altering the content and conceptual structure of the output text. Consequently, the algorithm may exhibit suboptimal performance within certain generation contexts where the content or conceptual direction is explicitly specified or mandated. 
% This observation highlights the nuanced nature of the algorithm's application, suggesting the need for selective deployment depending on the specific requirements and constraints of the task at hand.
% In Table \ref{tab:case study ROC} and \ref{tab:case study AASC}, we present example outputs from fine-tuned LLaMA2, GPT-2 large, and T5 large pre-trained language models (PLMs) on ROCStories and AASC datasets. The results demonstrate that our method effectively guides the models to generate more original sentences, showcasing the positive impact of our approach on enhancing content originality.

\begin{table*}[t]
% \resizebox{0.98\textwidth}{!}{
\begin{tabular}{ccp{10.9cm}}
\toprule
\multirow{7}{*}{LLaMA2}  & \multirow{3}{*}{Input}          & I was sick of being unemployed. I decided it was time to get a job. I scoured listings on the internet and sent out resumes. Finally, a company asked me to come in for an interview.                                      \\ \cline{2-3} 
                            & \multirow{2}{*}{Default Output} & I was so happy. When I got there, the receptionist gave me a clipboard with a bunch of paperwork on it and \textcolor{red}{said I had to} fill it out                                                                                    \\ \cline{2-3} 
                            & \multirow{2}{*}{SPCD Output}     & I was thrilled. I prepared for the meeting. I made sure to have my hair and clothes looking great. I had to look my best to get the job.
                                 \\ \midrule
\multirow{8}{*}{GPT-2}  & \multirow{3}{*}{Input}           & Thomas lost his job three weeks ago. He had no idea how he was going to pay his rent. His friend told him about a website that paid five dollars per blog. Thomas didn't think he could raise five hundred dollars there. \\ \cmidrule(l){2-3} 
                               & \multirow{2}{*}{Default Output} &  \textcolor{red}{He asked his friends what they thought} would happen. \textcolor{red}{They told him that they} were going to hire him.                                                                                                          \\ \cmidrule(l){2-3} 
                               & \multirow{2}{*}{SPCD Output}     & He got five hundred dollars together and started the website. He now has a website and is paid to write about things he knows!                                                                                             \\ \midrule
\multirow{6}{*}{T5}     & \multirow{2}{*}{Input}           & Tim was fishing on the bank. He felt a bit on his line. He yanked up real hard. The fish on his line flew out of the water.                                                                                                 \\ \cmidrule(l){2-3} 
                               & \multirow{2}{*}{Default Output} & Tim \textcolor{red}{let go of his} hook. \textcolor{red}{He was happy he was finally able to reel in his}  fish.                                                                         \\ \cmidrule(l){2-3} 
                               & SPCD Output    & He yanked it again, this time a lot harder. The line snapped in half.        \\ \bottomrule
\end{tabular}
% }
\caption{Example outputs from fine-tuned LLaMA2, GPT-2 large and T5 large PLMs on ROCStories. Text segments in \textcolor{red}{red} indicate plagiarism fragments identified by GOT metric. Results suggest that our proposed SP contrastive decoding strategy enhances the originality of text generated by PLMs, effectively mitigating plagiarism.}
\label{tab:case study ROC}
\end{table*}

\begin{table*}[t]
% \scalebox{1}{
\begin{tabular}{ccp{10.9cm}}
\toprule
\multirow{8}{*}{LLaMA2} & \multirow{2}{*}{Input}          & This paper deals with predictions which extend to the next several words in the text.                                                                                                                                                 \\ \cmidrule(l){2-3}
                            & \multirow{3}{*}{Default Output} & The model is \textcolor{red}{based on an artificial neural network with a hidden layer} of sigmoidal  units. The network has two outputs, one of which predicts the next character and...                                                \\ \cmidrule(l){2-3}
                            & \multirow{2}{*}{SPCD Output}     & Predictive text is the technology behind the texting feature in many cellphones and other smartphones.
                                                       \\ \midrule
\multirow{8}{*}{GPT-2} & \multirow{2}{*}{Input}          & If the classified user tweets achieved an accuracy above a threshold (0.7), the user model was applied on the authored                                                                                                  \\ \cmidrule(l){2-3} 
                               & \multirow{3}{*}{Default Output} & users in the test set with equal weight. To assess the generalization \textcolor{red}{capability of our method} we performed \textcolor{red}{additional experiments in which we used} a single group                                                \\ \cmidrule(l){2-3} 
                               & \multirow{2}{*}{SPCD Output}     & user tweets. The user tweets classified with greater accuracy were then sent to the  corresponding authored user tweets                                                                                                    \\ \midrule
\multirow{8}{*}{T5}       & \multirow{2}{*}{Input}          & First we computed baseline scores using the approaches described in Section 3.2. All resulting scores are listed in Table 2.                                                                                              \\ \cmidrule(l){2-3} 
                               & \multirow{3}{*}{Default Output} & We also included baselines computed \textcolor{red}{using the same approach} as in the English subtask. We \textcolor{red}{also included the results obtained when using} the baselines                                                            \\ \cmidrule(l){2-3} 
                               & \multirow{2}{*}{SPCD Output}     & The first row of Table 2 shows the baseline scores obtained using the naive approach (Naive).                                                                                                                             \\ \bottomrule
\end{tabular}
% }
\caption{Example outputs from fine-tuned LLaMA2, GPT-2 large and T5 large PLMs on AASC. Text segments in \textcolor{red}{red} indicate plagiarism fragments identified by GOT metric. Results suggest that our proposed SP contrastive decoding strategy enhances the originality of text generated by PLMs, effectively mitigating plagiarism.}
\label{tab:case study AASC}
\end{table*}

% This is a section in the appendix.

\end{document}